\documentclass[final]{cvpr}

\usepackage{times}
\usepackage{epsfig}
\usepackage{graphicx}
\usepackage{amsmath}
\usepackage{amssymb}

\usepackage{comment}
\usepackage{color}

\usepackage{bm}
\usepackage{multirow}
\usepackage{booktabs}
\usepackage{graphicx}
\usepackage{wrapfig}
\usepackage{enumerate}
\usepackage{subfigure}
\usepackage{algorithm}
\usepackage{algpseudocode}
\usepackage{bbding}

\usepackage[pagebackref=true,breaklinks=true,colorlinks,bookmarks=false]{hyperref}
\usepackage[numbers, sort]{natbib}

\def\cvprPaperID{2036} 
\def\confYear{CVPR 2021}

\begin{document}

\title{Learning Probabilistic Ordinal Embeddings for Uncertainty-Aware Regression}


\author{Wanhua Li$^{1,2}$, Xiaoke Huang$^{1,2,3}$, Jiwen Lu$^{1,2,\ast}$, Jianjiang Feng$^{1,2}$, Jie Zhou$^{1,2}$\\
$^{1}$Department of Automation, Tsinghua University, China \\
$^{2}$Beijing National Research Center for Information Science and Technology, China \\
$^{3}$School of Artificial Intelligence, Beijing Normal University\\
{\tt\small li-wh17@mails.tsinghua.edu.cn }\\
{\tt \small xiaokehuang@mail.bnu.edu.cn \qquad \{lujiwen,jfeng,jzhou\}@tsinghua.edu.cn}
}

\maketitle

\let\thefootnote\relax\footnotetext{$^{\ast}$ Corresponding author}
\begin{abstract}
   Uncertainty is the only certainty there is. Modeling data uncertainty is essential for regression, especially in unconstrained settings. Traditionally the direct regression formulation is considered and the uncertainty is modeled by modifying the output space to a certain family of probabilistic distributions. On the other hand, classification based regression and ranking based solutions are more popular in practice while the direct regression methods suffer from the limited performance. How to model the uncertainty within the present-day technologies for regression remains an open issue. In this paper, we propose to learn probabilistic ordinal embeddings which represent each data as a multivariate Gaussian distribution rather than a deterministic point in the latent space. An ordinal distribution constraint is proposed to exploit the ordinal nature of regression. Our probabilistic ordinal embeddings can be integrated into popular regression approaches and empower them with the ability of uncertainty estimation.  Experimental results show that our approach achieves competitive performance. Code is available at \url{https://github.com/Li-Wanhua/POEs}.

\end{abstract}

\section{Introduction}

Regression, as a fundamental machine learning problem, requires to predict a continuous target value $y$ for a given data $\bm{x}$. It is also extensively studied
due to many important applications, such as age estimation~\cite{pan2018mean,li2019bridgenet,wen2020adaptive}, historical image dating~\cite{palermo2012dating,martin2014dating}, and image aesthetic assessment~\cite{kong2016photo,lee2019image,pan2019image}. Recent years have witnessed the enormous success~\cite{ren2016faster,li2020grapheccv,li2020graphicme} of Deep Neural Networks (DNNs). Therefore, dominant approaches~\cite{wen2020adaptive,kuhnke2019deep,pan2019image} for regression employ DNNs to leverage the powerful feature representations. As the most natural choice for regression, the \emph{direct regression} method predicts a scalar value $y$ given an input $\bm{x}$ and the model is trained with $L^1$ or $L^2$ loss.

Modeling data uncertainty is quite important for regression since it gives a clear probabilistic interpretation of the predictions. In fact, humans usually associate confidence in their judgments. For example, if a person is required to estimate the age of a highly blurred facial image, he will give an estimation with high uncertainty. However, DNNs do not necessarily grasp the data uncertainty which captures the noise inherent in the data~\cite{liu2019probabilistic}. Many recent works~\cite{kendall2017uncertainties,gast2018lightweight,liu2019probabilistic,varamesh2020mixture} have been proposed to model the uncertainty, which consider the \emph{direct regression} solutions and replace the point estimation with a probabilistic distribution $p(y | \bm{x})$ in the output space. For simplicity, they usually consider a certain family of distributions, such as Gaussian or Laplace, which naturally limits the expressiveness.

\begin{figure}[t]
  \centering
  \includegraphics[width=1.0\linewidth]{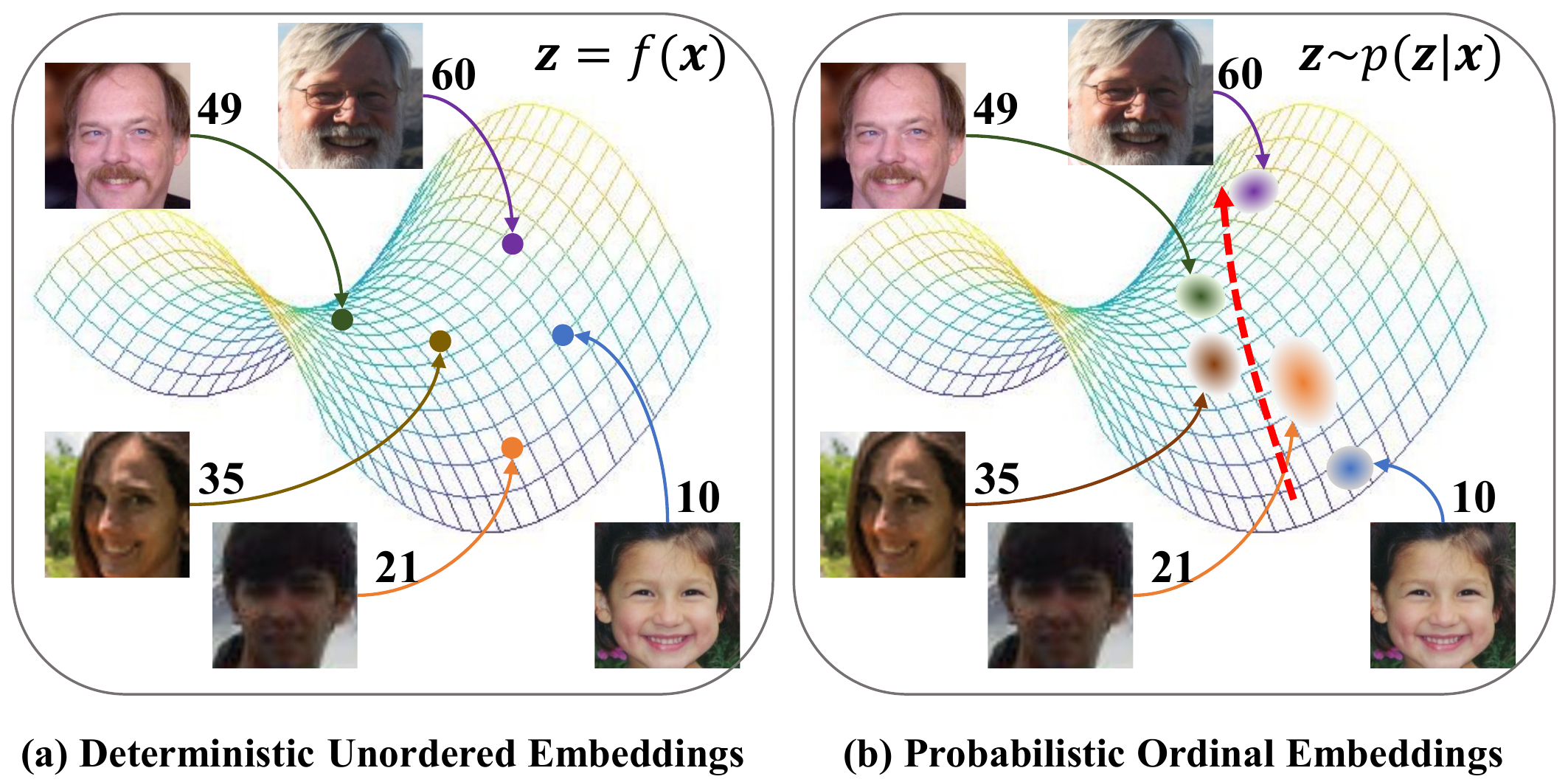}
  \caption{The key difference between deterministic unordered embeddings and probabilistic ordinal embeddings. We consider the problem of age estimation and display age labels next to the images. Deterministic unordered embeddings represent each facial image as a point in the latent space without considering the data uncertainty and the ordinal property. By contrast, probabilistic ordinal embeddings learn a distribution following the ordinal constraint in the embedding space. We see that the highly blurred image exhibits high uncertainty and the ordinal relations are preserved with our probabilistic ordinal embeddings.}
  \label{fig:motivation}
  \vspace {-0.6cm}
\end{figure}
%

Meanwhile, many advanced methods~\cite{pan2018mean,yang2018ssr,yang2019fsa} have been presented for regression, which significantly surpass the \emph{direct regression} approach. Some of the dominant approaches include classification based methods~\cite{rothe2018deep,ruiz2018fine} and ranking based methods~\cite{chang2011ordinal,niu2016ordinal,chen2017using}. Classification based method treats the regression problem as a multi-class classification problem by discretizing the target space into $C$ classes. Ranking based methods transform the regression problem into multiple binary classification sub-problems, where each binary classifier aims to predict whether the rank of a sample is larger than a specific value. Modeling data uncertainty within these popular regression methods is quite challenging. Unlike the \emph{direct regression} method which outputs a scalar value, both classification based regression and ranking based approaches have multiple neurons in the output layer. These neurons are highly correlated so it is difficult to choose a suitable distribution to replace them, which makes the above uncertainty-aware strategy unsuitable for these advanced regression methods.

To model the data uncertainty within the present-day regression methods, we propose Probabilistic Ordinal Embeddings (POEs), which estimate a multivariate Gaussian distribution, instead of a fixed point in the latent space. Most existing methods represent each input data as a deterministic point in the embedding space, \emph{i,e.}, $\bm{z} = f_{\theta}(\bm{x})$ is a point in $\mathbb{R}^D$, where $\bm{z} \in \mathbb{R}^D$ is the embedding, $f$ is the mapping function and $\theta$ is the parameter of $f$. However, it is difficult to give an accurate point embedding for noisy data, in which larger uncertainty should be exhibited in the embedding space~\cite{shi2019probabilistic,chang2020data}. We consider modeling the data uncertainty in the embedding space and representing each embedding as a random variable: $\bm{z} \sim p(\bm{z}|\bm{x}) \in \mathbb{R}^D$. Meanwhile, the ordinal relation in the target space should be preserved in
the embedding space since the uncertainty mainly spreads the probability mass across neighbors.
Therefore, an ordinal distribution constraint is further proposed to enforce the ordinal property.
Figure \ref{fig:motivation} shows the key difference between deterministic unordered embeddings and probabilistic ordinal embeddings.


In summary, the main contributions of our paper are threefold: (1)  To model the data uncertainty of regression, we propose probabilistic ordinal embeddings, which represent each data as a distribution following the ordinal constraint in the latent space.
(2)   Our method empowers the present-day regression methods with the ability of uncertainty estimation. A per-exemplar uncertainty score can be used to measure the reliability of predictions.
(3)   Comprehensive experiments illustrate that uncertainty modeling improves the performance and achieves very competitive performance on three real-world visual tasks.

\section{Related Work}
\textbf{Uncertainty Learning:} DNNs tend to give overconfident predictions~\cite{lakshminarayanan2017simple,kong2020sde} and learning to quantify the predictive uncertainty is essential in many real-world applications~\cite{gal2016dropout,kendall2018multi,ovadia2019can,van2020uncertainty}. Uncertainty learning for computer vision tasks can be categorized into regression settings such as age estimation~\cite{liu2019probabilistic}, and classification settings such as semantic segmentation~\cite{huang2018efficient}. There are two major types of uncertainty: \emph{epistemic (model) uncertainty} and \emph{aleatoric (data) uncertainty}~\cite{kendall2017uncertainties}.
Data uncertainty refers to the noise inherent in the input data which cannot be eliminated with more training data. Our work mainly focuses on data uncertainty. It is usually modeled by placing a distribution over a model's output~\cite{kendall2017uncertainties}. Specifically, the direct regression formulation is considered for regression and the scalar output is corrupted with Gaussian random noise. However, it is not suitable for present-day regression methods which usually have multiple neurons in the output layers~\cite{rothe2018deep,niu2016ordinal,chen2017using}. Placing a multivariate Gaussian distribution over the vector in output space cannot model the complex relations among the elements in the output vector. Recently, some methods~\cite{oh2018modeling,shi2019probabilistic,chang2020data} have proposed stochastic embeddings to model the data uncertainty for classification problems. However, these methods are not suitable for regression problems since they fail to model the ordinal nature of regression. In this paper, we propose to learn a distribution following the ordinal constraint in the latent space.

\textbf{Regression Methods:} Supervised regression aims to learn a mapping from the input $\bm{x} \in \mathcal{X}$ to a continuous target value $y \in \mathcal{Y}$. A variety of methods~\cite{pan2018mean,li2019bridgenet,ruiz2018fine,palermo2012dating,kong2016photo} have been proposed for regression in many computer vision tasks. Here we describe some popular methods. The most natural solution is direct regression, which trains a DNN with a single output neuron to predict the target value. However, the direct regression methods suffer from limited performance and are not very popular in practice~\cite{chen2017using,ruiz2018fine}. Recently, some classification based methods~\cite{yang2019fsa,ruiz2018fine} have been proposed, which divide the output space into several bins and treat each bin as an independent class. Some variants of this method, such as mean-variance loss~\cite{pan2018mean} and label distribution learning~\cite{geng2013facial,gao2018age}, have achieved great success in many fields~\cite{su2019soft,he2019s2gan,chen2020label}. Another dominant approach is the ranking based method~\cite{niu2016ordinal,liu2017deep}, which constructs multiple binary classifiers and trains them with rank labels. An ordinal constraint was used in~\cite{herbrich2000large}, but it failed to incorporate the data uncertainty on the embedding space.
More recent works formulate the regression problem from a probabilistic perspective~\cite{kendall2017uncertainties,gustafsson2020energy}. For example, He \emph{et al.}~\cite{he2019bounding} placed a Gaussian distribution over the output to model the uncertainty. Mixture Dense Regression~\cite{varamesh2020mixture} further replaced the output with a mixture of Gaussian. While these methods are only compatible with the direct regression methods, we are committed to modeling data uncertainty within dominant regression approaches.

\begin{figure*}[t]
\begin{center}
   \includegraphics[width=1.0\linewidth]{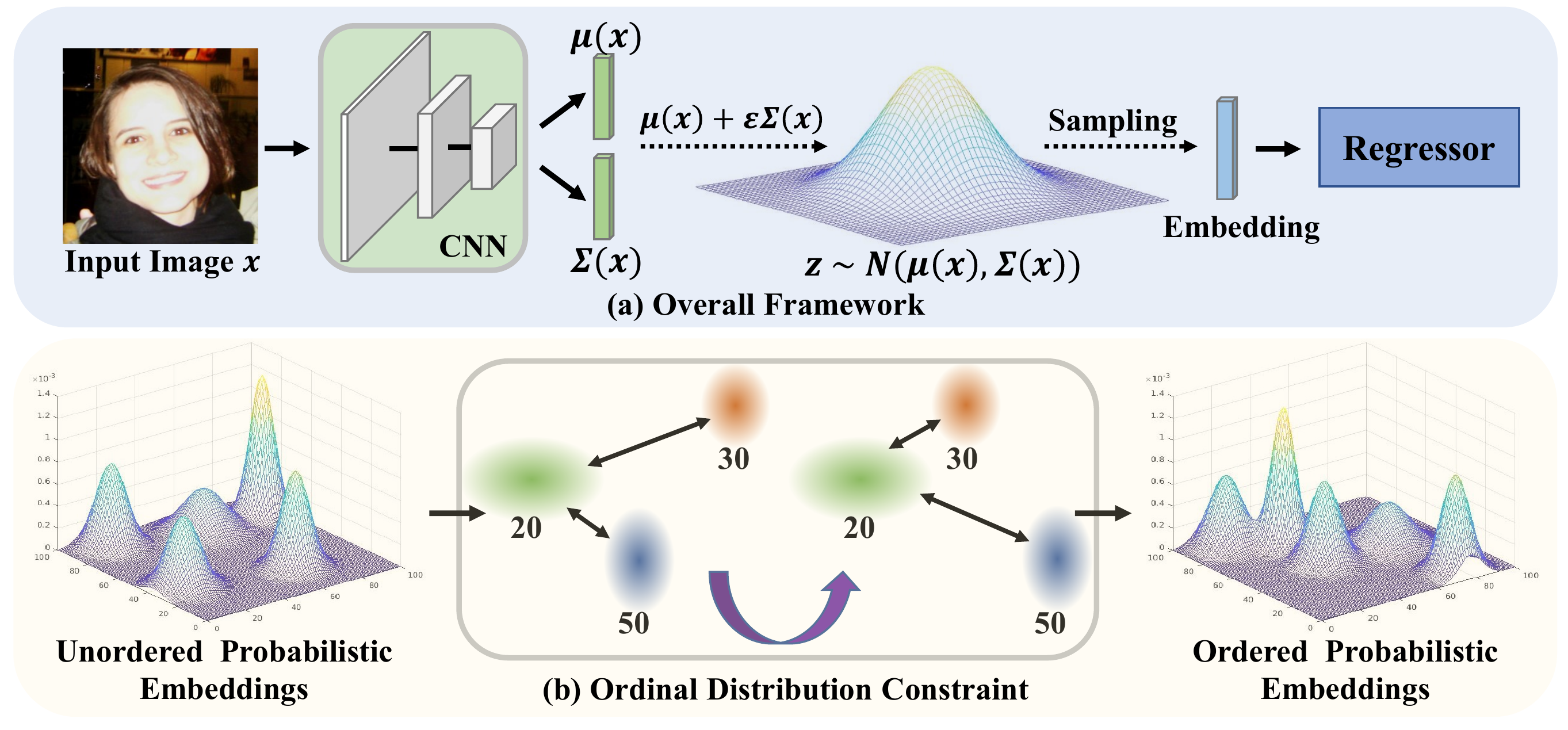}
\end{center}
\vspace{-0.5cm}
   \caption{(a) The overall framework of our approach. Given an input image $\bm{x}$, we first obtain the parameters $\bm{\mu}(\bm{x})$ and $\bm{\Sigma}(\bm{x})$, which correspond to the mean and diagonal covariance of a multivariate Gaussian distribution respectively. Then we sample embeddings from this distribution with the reparameterization  trick. Lastly, the sampled embeddings are sent to a popular regressor for prediction. (b) The idea of ordinal distribution constraint. For three samples with labels of 20, 30, and 50, we enforce the ordinal relations in the embedding space with the symmetrized Kullback-Leibler divergence or 2-Wasserstein distance. In the end, we expect that ordered probabilistic embeddings are obtained by applying the ordinal distribution constraint in the latent space.}
\label{fig:flowchart}
\vspace{-0.5cm}
\end{figure*}

\section{Proposed Approach}
In this section, we first revisit the formulations of popular regression methods with deterministic embeddings. Then we introduce the details of probabilistic embeddings. Lastly, we present the ordinal distribution constraint and show how to integrate the proposed probabilistic ordinal embeddings into different regression methods.

\subsection{Preliminaries}

\textbf{Direct Regression Solution:} Direct regression method employs learnable parameters $\bm{w} \in \mathbb{R}^{D \times 1}$ to project the point embedding $\bm{z} = f_{\theta}(\bm{x})$ into a scalar value $\hat{y}$:
\vspace{-0.2cm}
\begin{equation}
\hat{y} =  \bm{w}^\top f_{\theta}(\bm{x}),
\label{equ:pre:DR}
\vspace{-0.2cm}
\end{equation}
where $\cdot ^\top$ denotes transposition. Then the model is usually trained with $L^1$ or $L^2$ loss. We present the formulation with $L^2$ loss as follows:
\vspace{-0.2cm}
\begin{equation}
\mathcal{L}_{DR}(\bm{x}) =  \left \| y - \bm{w}^\top f_{\theta}(\bm{x}) \right \| ^2.
\label{equLoss:DR}
\vspace{-0.2cm}
\end{equation}

\textbf{Regression by Classification:} Many methods discretizing the target space into $C$ classes and treat regression as a classification problem. We obtain the predictions with a classification layer parameterized by $\bm{W} = [\bm{w}_1, \bm{w}_2, ..., \bm{w}_{C}]^\top \in \mathbb{R}^{K \times D}$. Cross entropy loss is utilized to optimize the networks. Specifically, for a sample $\bm{x}$ with the ground truth $c$, the loss is defined as follows:
\begin{equation}
\begin{aligned}
\mathcal{L}_{Cls}(\bm{x}) &= -\log p(y = c|\bm{x};\theta, \bm{W}) \\
&= -\log \frac{\exp(\bm{w}_c^\top f_{\theta}(\bm{x}))}{\sum_{r=1}^{C} \exp(\bm{w}_r^\top f_{\theta}(\bm{x}))}.
\end{aligned}
\label{equ:Loss:Cla}
\end{equation}

Several strategies can be employed to make predictions from the classification results~\cite{rothe2018deep}. A simple choice is to use the class with the highest probability:
\begin{equation}
\hat{y} =  \mathop{\arg\max}_{r \in \{1,2,...,C\}}  p(y = r|\bm{x};\theta, \bm{W}).
\label{equ:pre:Cla}
\end{equation}

\textbf{Ranking based Regression:} Classification based methods treat different classes independently, which ignores the ordinal relations of different classes. Ranking based methods construct $C - 1$ binary classifiers for $C$ classes, where $k$-th one predicts whether the rank of the sample $\bm{x}$ is larger than $k$. The $C - 1$ labels are constructed as $r_1, r_2, ..., r_{C-1} \in \{0,1\}$ for a given input $\bm{x}$. Then all $C - 1$ classifiers are jointly trained with the cross-entropy loss:
\begin{equation}
\begin{aligned}
\mathcal{L}_{Rank}(\bm{x}) &= - \sum_{k=1}^{C-1} \log p(b_k = r_k|\bm{x};\theta, \bm{W}_k) \\
&= - \sum_{k=1}^{C-1} \log  \frac{\exp(\bm{w}_{k,r_k}^\top f_{\theta}(\bm{x}))}{\sum_{i=0}^{1} \exp(\bm{w}_{k,i}^\top f_{\theta}(\bm{x}))},
\end{aligned}
\label{equ:Loss:rank}
\end{equation}
where $b_k$ is the output of the $k$-th binary classifier, and $\bm{W}_k = [\bm{w}_{k,0}, \bm{w}_{k,1}]^\top \in \mathbb{R}^{2 \times D}$ is the parameters of the $k$-th binary classifier. The decoding strategy is defined as:
\begin{equation}
\hat{y} = 1 + \sum_{k=1}^{C-1} b_k.
\label{equ:pre:rank}
\end{equation}

\subsection{Probabilistic Embeddings for Regression}
To model the data uncertainty for different regression methods, we consider treating embeddings as probabilistic distributions, \emph{i.e.,} $\bm{z} \sim p(\bm{z}|\bm{x})$. Given an input $\bm{x}$, we can obtain a probabilistic prediction by marginalizing over the embedding distribution:
\vspace{-0.2cm}
\begin{equation}
p(y|\bm{x}) = \int p(y|\bm{z})p(\bm{z}|\bm{x})  {\rm d}\bm{z}.
\label{equ:bayes}
\vspace{-0.2cm}
\end{equation}

It is difficult to give a closed-form expression for the integral in \eqref{equ:bayes}. Therefore, we perform Monte-Carlo sampling from $\bm{z}^{(t)} \sim p(\bm{z}|\bm{x})$ to approximate it:
\vspace{-0.2cm}
\begin{equation}
p(y|\bm{x}) \approx  \frac{1}{T} \sum_{t=1}^T p(y|\bm{z}^{(t)}),
\label{equ:sampling}
\vspace{-0.2cm}
\end{equation}
where $T$ is the number of samples. We set $T$ to 50 in our experiments for a good trade-off.

To sample from the distribution $p(\bm{z}|\bm{x})$, we define the embedding $\bm{z}$ as a multivariate Gaussian distribution for simplicity:
\vspace{-0.2cm}
\begin{equation}
p(\bm{z}|\bm{x}) = \mathcal{N}(\bm{z};\bm{\mu}(\bm{x}), \bm{\Sigma}(\bm{x})),
\label{equ:embedding}
\vspace{-0.2cm}
\end{equation}
where the mean $\bm{\mu}(\bm{x})$ and the diagonal covariance $\bm{\Sigma}(\bm{x})$ are input-dependent: $\bm{\mu}(\bm{x}) = f_{\theta_1}(\bm{x}), {\rm diag} (\bm{\Sigma}(\bm{x})) = f_{\theta_2}(\bm{x})$, and $\theta_1$ and $\theta_2$ refer to the corresponding model parameters. We implement $f_{\theta_1}(\bm{x})$ and $f_{\theta_2}(\bm{x})$ with a shared backbone and two head branches. The mean $\bm{\mu}(\bm{x})$ can be regarded as the most likely embedding values while the diagonal covariance $\bm{\Sigma}(\bm{x})$ can be interpreted as the data uncertainty. Naturally, larger variance means higher uncertainty. While sampling operation is not differentiable, we consider the reparameterization trick~\cite{kingma2014auto} to enable backpropagation:
\begin{equation}
\bm{z}^{(t)} = \bm{\mu}(\bm{x}) + {\rm diag} (\sqrt{\bm{\Sigma}(\bm{x})}) \cdot \bm{\epsilon}^{(t)}, \bm{\epsilon}^{(t)} \sim \mathcal{N}(\bm{0},\bm{I}).
\label{equ:reparam}
\end{equation}
Therefore, we first sample noise from $\mathcal{N}(\bm{0},\bm{I})$ and then obtain $\bm{z}$ following \eqref{equ:reparam} instead of directly sampling from $\mathcal{N}(\bm{\mu}(\bm{x}), \bm{\Sigma}(\bm{x}))$.

To integrate our method into existing regression approaches, we treat the $p(y|\bm{z})$ as the estimator with different formulations and rewrite the loss functions accordingly. For the direct regression solution, our method minimizes the following loss:
\begin{equation}
\vspace{-0.2cm}
\mathcal{L}_{PDR}(\bm{x}) = \frac{1}{T} \sum_{t=1}^{T}\left\|y - \bm{w}^\top \bm{z}^{(t)} \right\|^2
\label{equ:Loss:PDR}
\vspace{-0.2cm}
\end{equation}

For classification based regression, the objective function is defined as follows:
\begin{equation}
\begin{aligned}
\mathcal{L}_{PCls}(\bm{x}) &= - \frac{1}{T} \sum_{t=1}^{T} \log p(y = c|\bm{z}^{(t)};\theta_1,\theta_2, \bm{W}) \\
&= - \frac{1}{T} \sum_{t=1}^{T} \log \frac{\exp(\bm{w}_c^\top \bm{z}^{(t)})}{\sum_{r=1}^{C} \exp(\bm{w}_r^\top \bm{z}^{(t)})}.
\end{aligned}
\label{equ:Loss:PCls}
\end{equation}

Furthermore, the loss function for ranking based methods is given by:
\begin{equation}
\begin{aligned}
\mathcal{L}_{PRank}(\bm{x}) &= - \frac{1}{T} \sum_{t=1}^{T} \sum_{k=1}^{C-1} \log p(b_k = r_k|\bm{z}^{(t)};\theta_1,\theta_2, \bm{W}_k) \\
&=  - \frac{1}{T} \sum_{t=1}^{T} \sum_{k=1}^{C-1} \log  \frac{\exp(\bm{w}_{k,r_k}^\top \bm{z}^{(t)})}{\sum_{i=0}^{1} \exp(\bm{w}_{k,i}^\top \bm{z}^{(t)})}.
\end{aligned}
\label{equ:Loss:rank}
\end{equation}

In this way, we can train the present-day regression methods with probabilistic embeddings. We show the above pipeline in Figure \ref{fig:flowchart}(a).
Note that the Monte-Carlo sampling is ONLY used during training since we observe no significant performance difference during inference.

\subsection{Ordinal Distribution Constraint}
The target space of regression is continuous and ordinal, which means that the uncertainty mainly spreads probability mass across the locations that  correspond to the neighbors of ground truth value. For example, if we estimate the age of a person as 30, then his uncertainty in the embedding space depicts that he is around 30. Therefore, we expect that the ordinal relation in the target space can be preserved in the embedding space. We consider applying a set of ordinal distribution constraints to enforce the ordinal property in the embedding space. For a given triplet $(\bm{x}_l,\bm{x}_m,\bm{x}_n)$, and its associated label $(y_l,y_m,y_n)$ and probabilistic embedding $(\bm{z}_l,\bm{z}_m,\bm{z}_n)$, we aim to preserve the following constraint:
\begin{equation}
 d(\bm{z}_l,\bm{z}_m) < d(\bm{z}_l,\bm{z}_n) \Leftrightarrow   \left| y_l - y_m \right| < \left| y_l - y_n \right|,
\label{equ:loss:ordinal}
\end{equation}
where $d(\cdot)$ is a distance metric. Note that the triplet is not the conventional triplet since the positive and negative samples are not fixed
for a given anchor sample. More details about our triplet are provided in the supplementary material.

We let $\mathcal{S} = \{(l,m,n)\big| \left| y_l - y_m \right| < \left| y_l - y_n \right|\}$, then the ordinal distribution constraint is formulated with hinge loss:
\begin{equation}
\mathcal{L}_{Ord} = \frac{1}{|\mathcal{S}|} \sum_{(l,m,n) \in \mathcal{S}} \max (0, d(\bm{z}_l,\bm{z}_m)  + \delta -  d(\bm{z}_l,\bm{z}_n) ),
\label{equ:hinge}
\end{equation}
where $\delta$ is the margin. Note that our ordinal distribution constraint in \eqref{equ:hinge} is non-trivial, given that metric $d(\cdot)$ measures the distance between two distributions rather than that between two points. Therefore, some common choices for $d(\cdot)$, such as the Euclidean distance, are not suitable.

Kullback-Leibler divergence is commonly used as the metric to measure the difference between two distributions in many applications, such as label distribution learning~\cite{gao2017deep,shen2017label,gao2018age}. However, it is an asymmetric measure and does not satisfy the triangle inequality. In this paper, we consider a symmetrized Kullback-Leibler divergence $SKL\left(P, Q\right) = KL\left(P \middle\| Q\right) + KL\left(Q \middle\| P\right)$ and ignore the triangle inequality issue. Furthermore, We can obtain the closed-form expression for symmetrized Kullback-Leibler divergence between two multivariate Gaussian distributions. For example, the symmetrized KL divergence between $\bm{z}_l$ and $\bm{z}_m$ is given by:
\begin{equation}
\begin{aligned}
SKL(\bm{z}_l, \bm{z}_m) &= KL\left(\bm{z}_l \middle\| \bm{z}_m\right) + KL\left(\bm{z}_m \middle\| \bm{z}_l\right) \\
&= \frac{1}{2} \sum_{j=1}^{D} [\frac{(\bm{\sigma}_l^j)^2}{(\bm{\sigma}_m^j)^2} + \frac{(\bm{\sigma}_m^j)^2}{(\bm{\sigma}_l^j)^2} -2 \\
&+ \frac{(\bm{\mu}_l^j- \bm{\mu}_m^j)^2}{(\bm{\sigma}_l^j)^2} + \frac{(\bm{\mu}_l^j- \bm{\mu}_m^j)^2}{(\bm{\sigma}_m^j)^2}],
\end{aligned}
\label{equ:SKL}
\end{equation}
where $\bm{\mu}_l^j,\bm{\mu}_m^j,\bm{\sigma}_l^j$, and $\bm{\sigma}_m^j$ are the $j$-th dimension of $\bm{\mu}(\bm{x}_l),\bm{\mu}(\bm{x}_m),{\rm diag}(\sqrt{\bm{\Sigma}(\bm{x}_l)})$, and ${\rm diag}(\sqrt{\bm{\Sigma}(\bm{x}_m)})$ respectively. We can easily obtain $SKL(\bm{z}_l, \bm{z}_n)$ following a similar formulation.

On the other hand, Wasserstein distance has attracted increasing attention due to its success in generative adversarial networks~\cite{arjovsky2017towards,arjovsky2017wasserstein,gulrajani2017improved}. Wasserstein distance can be viewed as the minimum cost of turning one pile into the other. In this paper, we also consider utilizing the Wasserstein metric to measure the distance of two Gaussian distributions. Specifically, the 2-Wasserstein distance is employed since it gives a closed-form solution for two normal distributions:
\begin{equation}
\begin{aligned}
W_2(\bm{z}_l, \bm{z}_m)^2 &= W_2(\mathcal{N}(\bm{\mu}_l, \bm{\Sigma}_l),\mathcal{N}(\bm{\mu}_m, \bm{\Sigma}_m))^2\\
&= \sum_{j=1}^{D}(\bm{\mu}_l^j - \bm{\mu}_m^j)^2 + (\bm{\sigma}_l^j - \bm{\sigma}_m^j)^2.
\end{aligned}
\label{equ:Wdistance}
\end{equation}

With the formulations in \eqref{equ:SKL} and \eqref{equ:Wdistance}, we can easily calculate the ordinal distribution constraint loss in \eqref{equ:loss:ordinal} under different distance metrics. We further depict the idea of the ordinal distribution constraint in Figure \ref{fig:flowchart}(b).

Although the data uncertainty can be modeled with the introduced $\bm{\Sigma}(\bm{x})$, our model may predict small $\bm{\Sigma}(\bm{x})$ for all samples and approximately degenerate into deterministic embeddings. To address this issue, we borrow the idea of the variational information bottleneck~\cite{alemi2017deep} and introduce a regularization term. The information bottleneck principle aims to learn a latent encoding which captures the task-related parts of $\bm{x}$ while forgetting the irrelevant parts of the input. Concretely, we introduce a KL divergence term to constrain our probabilistic ordinal embeddings $\bm{z}$:
\begin{equation}
\begin{aligned}
\mathcal{L}_{VIB}(\bm{x}) &= KL\left( \mathcal{N}(\bm{\mu}(\bm{x}), \bm{\Sigma}(\bm{x})) \middle\| \mathcal{N}(\bm{0}, \bm{I})\right)  \\
&= \frac{1}{2}\sum_{j=1}^{D}((\bm{\mu}^j)^2 + (\bm{\sigma}^j)^2 - \log (\bm{\sigma}^j)^2 -1),\\
\end{aligned}
\label{equ:loss:VIB}
\end{equation}
where $\bm{\mu}^j$ and $\bm{\sigma}^j$ denote the $j$-th dimension of $\bm{\mu}(\bm{x})$ and ${\rm diag}(\sqrt{\bm{\Sigma}(\bm{x})})$ respectively.

In the end, we combine the above three ingredients to train our method. For example, we train the probabilistic ordinal embeddings for classification based regression with the following loss function:
\begin{equation}
\begin{aligned}
\mathcal{L}_{POE} = \frac{1}{\left| \mathcal{D} \right|} \sum_{\bm{x} \in \mathcal{D}} \mathcal{L}_{PCls}(\bm{x}) + \alpha \mathcal{L}_{Ord} +  \frac{\beta}{\left| \mathcal{D} \right|} \sum_{\bm{x} \in \mathcal{D}} \mathcal{L}_{VIB}(\bm{x}),
\end{aligned}
\label{equ:loss:POE}
\end{equation}
where $\mathcal{D}$ denotes the training set, $\alpha$ and $\beta$ are trade-off hyper-parameters which control the importance of three ingredients. We can easily infer the total loss functions for other present-day regression solutions.

\section{Experiments}
In this section, we conducted extensive experiments to validate the effectiveness of the proposed approach.

\subsection{Age Estimation}

\textbf{Datasets:} Age estimation attempts to predict the age for a given facial image. In our experiments, we apply our method to two widely used age datasets: the MORPH II~\cite{ricanek2006morph} dataset and the Adience~\cite{levi2015age} dataset. MORPH II dataset includes 55,134 facial images of 13,618 individuals. Each image is labeled with the exact age value ranging from 16 to 77 years. We employ the popular protocol as in \cite{rothe2018deep,shen2018deep,li2019bridgenet}, which selects 5,492 images from Caucasian descent to avoid the cross-race influence. Then these images are divided into two subsets: 80\% for training and 20\% for testing. Adience dataset contains 26,580 Flickr photos of 2,284 subjects. The age is annotated with eight groups: 0-2, 4-6, 8-13, 15-20, 25-32, 38-43, 48-53, and over 60 years old. For evaluation, we adopt the standard five-fold, subject-exclusive cross-validation protocol as used in ~\cite{liu2018constrained}.

\textbf{Experimental Settings:} For the MORPH II database, most existing methods~\cite{tan2018efficient,li2019bridgenet,wen2020adaptive} utilized a VGG-16 network that was pre-trained on the large-scale IMDB-WIKI dataset~\cite{rothe2018deep} as the feature extractor. For a fair comparison, the same strategy was adopted in our experiments. Meanwhile, the ImageNet pre-trained VGG-16 network was used for the Adience face database since most competing methods~\cite{liu2019probabilistic,diaz2019soft} used this model. We directly employed the penultimate fully connected layer of VGG-16 to predict the mean $\bm{\mu}(\bm{x})$ with embedding dimensions of 4096. An additional head branch was used to output the diagonal covariance $\bm{\Sigma}(\bm{x})$. We implemented it with the architecture: \texttt{FC-BN-exp}. All experiments in this paper employed this architecture to model the data uncertainty. For both age databases, we trained models for 50 epochs with a batch size of 32. For optimization,  Adam optimizer was utilized with a learning rate of 0.0001. For all experiments in this paper, we set the hyper-parameters $\alpha$ and $\beta$ to 0.0001 and 0.00001 respectively.
For the MORPH II database, we employed the MAE metric, which calculated the mean absolute error between ground truths and predictions. For the Adience face database, we also reported the classification accuracy for a more comprehensive comparison. To report the classification accuracy for the direct regression method and ranking based method, the closest class of the regression result was taken as the predicted class.

\begin{table}[tbp]
\caption{Results of symmetrized KL divergence with different margins on the MORPH II dataset.}
\vspace {0.2cm}
\label{table:ablation:SKL}
\centering
\begin{tabular}{lcccccc}
\toprule
Margin & 1 & 2 & 5 & 10 & 20 &50\\
\midrule
MAE & 2.38 & \textbf{2.35} & \textbf{2.35} & 2.37 & 2.41 &  2.41 \\
\bottomrule
\end{tabular}
\vspace {-0.2cm}
\end{table}

\begin{table}[tbp]
\caption{Results of  2-Wasserstein distance with different margins on the MORPH II dataset.}
\vspace {0.2cm}
\label{table:ablation:W2}
\centering
\begin{tabular}{lcccccc}
\toprule
Margin & 10 & 20 & 50 & 100 & 200 & 500\\
\midrule
MAE & 2.40 & 2.41 & 2.37 & \textbf{2.36} & 2.38 & 2.39\\
\bottomrule
\end{tabular}
\vspace {-0.2cm}
\end{table}

\begin{table}[tbp]
\caption{Ablation experiments on the MORPH II dataset.}
\vspace {0.2cm}
\renewcommand\tabcolsep{1.5pt}
\label{table:ablation:comp}
\centering
\begin{tabular}{l|ccccc|ccccc}
\toprule
\multirow{2}*{Choice} & \multicolumn{10}{c}{Regression Methods} \\
\cline{2-11}
~ & \multicolumn{5}{c|}{Classification} & \multicolumn{5}{c}{Ranking} \\
\hline
BL & \Checkmark &  \Checkmark&  \Checkmark & \Checkmark  &  \Checkmark & \Checkmark &  \Checkmark &   \Checkmark & \Checkmark &   \Checkmark \\
P-Emb & & \Checkmark &\Checkmark & \Checkmark  & \Checkmark  &  &  \Checkmark & \Checkmark &\Checkmark &  \Checkmark  \\
$\mathcal{L}_{Ord}$ & & &\Checkmark &  & \Checkmark &    &   &  \Checkmark &    & \Checkmark \\
$\mathcal{L}_{VIB}$ & & &  & \Checkmark  & \Checkmark&  &   &  & \Checkmark  & \Checkmark \\
\hline
MAE & 2.64 & 2.45 &  2.37 & 2.39 & 2.35 & 2.62  & 2.53 & 2.45 & 2.48 & 2.43 \\
\bottomrule
\end{tabular}
 \vspace {-0.5cm}
\end{table}

\textbf{Parameters Analysis:} We first investigate the influence of different distance metrics and different margins $\delta$ on the MORPH II database. Table \ref{table:ablation:SKL} shows the results with symmetrized KL divergence. We see that our method attains the best performance when the margin $\delta$ is set to 2 or 5. Furthermore, the performance varies slightly with different margins $\delta$, which shows our method is robust to the hyper-parameter $\delta$. We further list the results with 2-Wasserstein distance in Table \ref{table:ablation:W2}. We see that our method is also robust to the margin $\delta$ for 2-Wasserstein distance. Both metrics can achieve similar best results using appropriate margins, which demonstrates the robustness of our method to distance metrics. In the following experiments, we employ the symmetrized KL divergence with margin $\delta=5$.

\textbf{Ablation Study:} To validate the effectiveness of the proposed probabilistic ordinal embeddings, we conducted ablation studies on the MORPH II database with two popular regression methods. Table \ref{table:ablation:comp} shows the comparison results in different settings, where BL stands for baseline that learns unordered deterministic embeddings, and P-Emb means learning probabilistic embeddings. It is clear that the probabilistic embeddings boost the baseline methods, improving the MAE by 0.19 and 0.09 for the classification based method and ranking based method respectively. We find that the ordinal distribution constraint $\mathcal{L}_{Ord}$ and the regularization term $\mathcal{L}_{VIB}$ can further improve the performance while the improvement brought by the $\mathcal{L}_{Ord}$ is more significant. Note that the VIB loss is used to stabilize the training process rather than improve performance, so for the $\mathcal{L}_{Ord}$ method, further combining with $\mathcal{L}_{VIB}$ leads to marginal improvement. However, the $\mathcal{L}_{Ord}$ effectively outperforms the P-Emb method, showing the superiority of the proposed ordinal distribution constraint.

\begin{table}[t]
\caption{The comparisons between our method and other state-of-the-art methods on the MOPRH II dataset.}
\vspace {0.2cm}
\label{table:SOTA:MORPH}
\renewcommand\tabcolsep{10pt}
\centering
\begin{tabular}{lcc}
\toprule
Methods  & MAE & Year \\
\midrule
DRFs~\cite{shen2018deep} & 2.91 & 2018 \\
AGEn~\cite{tan2018efficient}& 2.52 & 2018 \\
BridgeNet~\cite{li2019bridgenet} & 2.38 & 2019 \\
AVDL~\cite{wen2020adaptive} & 2.37 & 2020 \\
\midrule
D-Regression & 2.74  & -   \\
D-Regression  + POE  & 2.50 & -   \\
Classification & 2.64 & -   \\
Classification + POE & \textbf{2.35} & -   \\
Ranking & 2.62 & -   \\
Ranking + POE & 2.43 & -  \\
\bottomrule
\end{tabular}
\vspace {-0.3cm}
\end{table}

\begin{table}[t]
\caption{Results on the Adience face dataset.}
\vspace {0.2cm}
\label{table:SOTA:Adience}
\centering
\begin{tabular}{lcc}
\toprule
Methods  & Accuracy(\%) & MAE  \\
\midrule
CNNPOR   \cite{liu2018constrained} & 57.4 $\pm$ 5.8 & 0.55 $\pm$ 0.08   \\
GP-DNNOR \cite{liu2019probabilistic} & 57.4 $\pm$ 5.5 & 0.54 $\pm$ 0.07  \\
SORD \cite{diaz2019soft} & 59.6 $\pm$ 3.6 & 0.49 $\pm$ 0.05  \\
\midrule
D-Regression & 56.3 $\pm$ 4.9 & 0.56 $\pm$ 0.07   \\
D-Regression  + POE  & 57.6 $\pm$ 4.2 & 0.54 $\pm$ 0.05   \\
Classification~\cite{liu2018constrained} & 54.0 $\pm$ 6.3 & 0.61 $\pm$ 0.08   \\
Classification + POE & \textbf{60.5 $\pm$ 4.4} & \textbf{0.47 $\pm$ 0.06}   \\
Ranking~\cite{niu2016ordinal} & 56.7 $\pm$ 6.0 & 0.54 $\pm$ 0.08   \\
Ranking + POE & 60.3 $\pm$ 4.4 & 0.48 $\pm$ 0.07  \\
\bottomrule
\end{tabular}
\vspace {-0.5cm}
\end{table}

\begin{figure*}[t]
  \centering
  \subfigure[MAE vs uncertainty on MORPH II dataset]{
    \label{fig:corre:morph2}
    \includegraphics[width=0.32\linewidth]{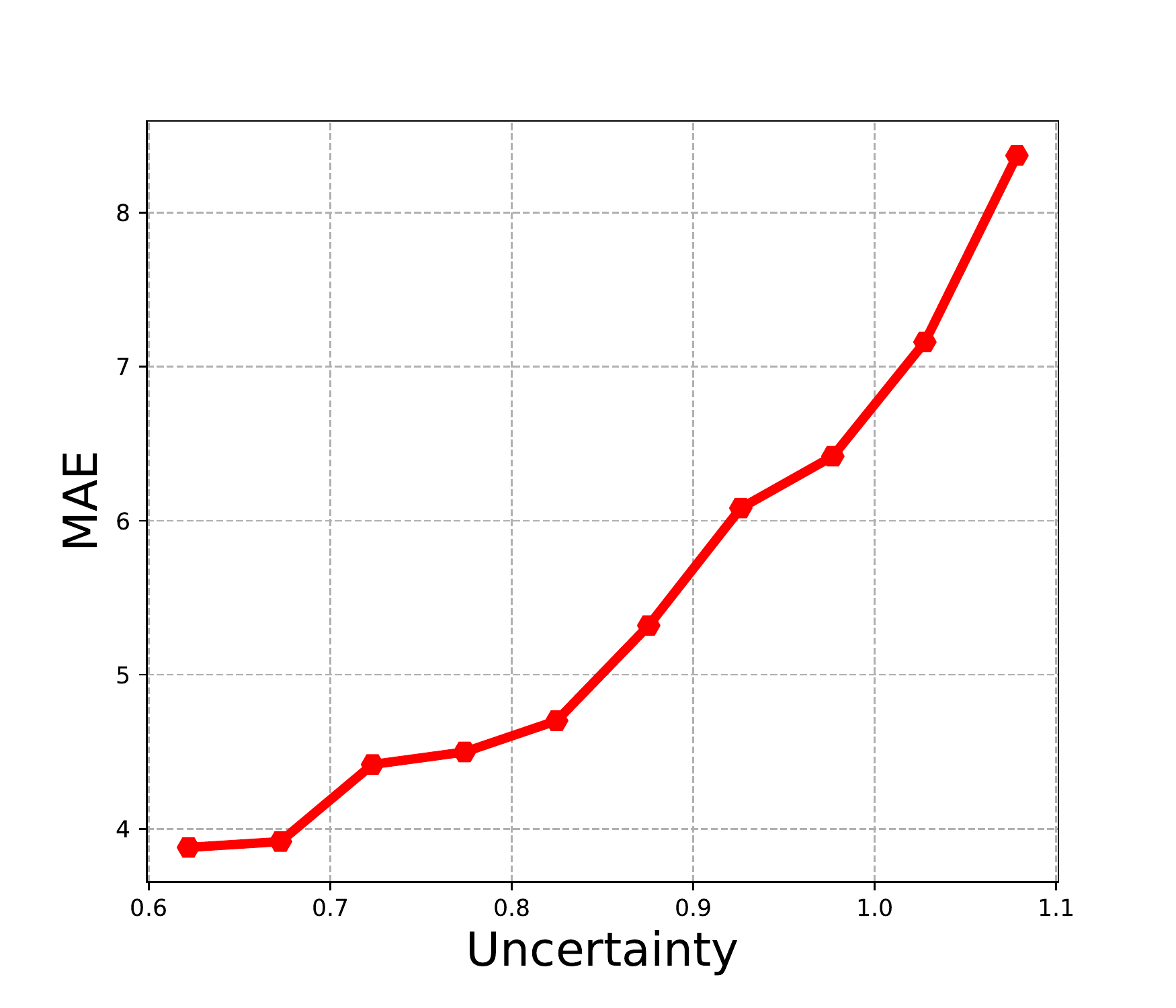}
  }
  \subfigure[Accuracy vs uncertainty on Adience dataset]{
    \label{fig:corre:adience1}
    \includegraphics[width=0.32\linewidth]{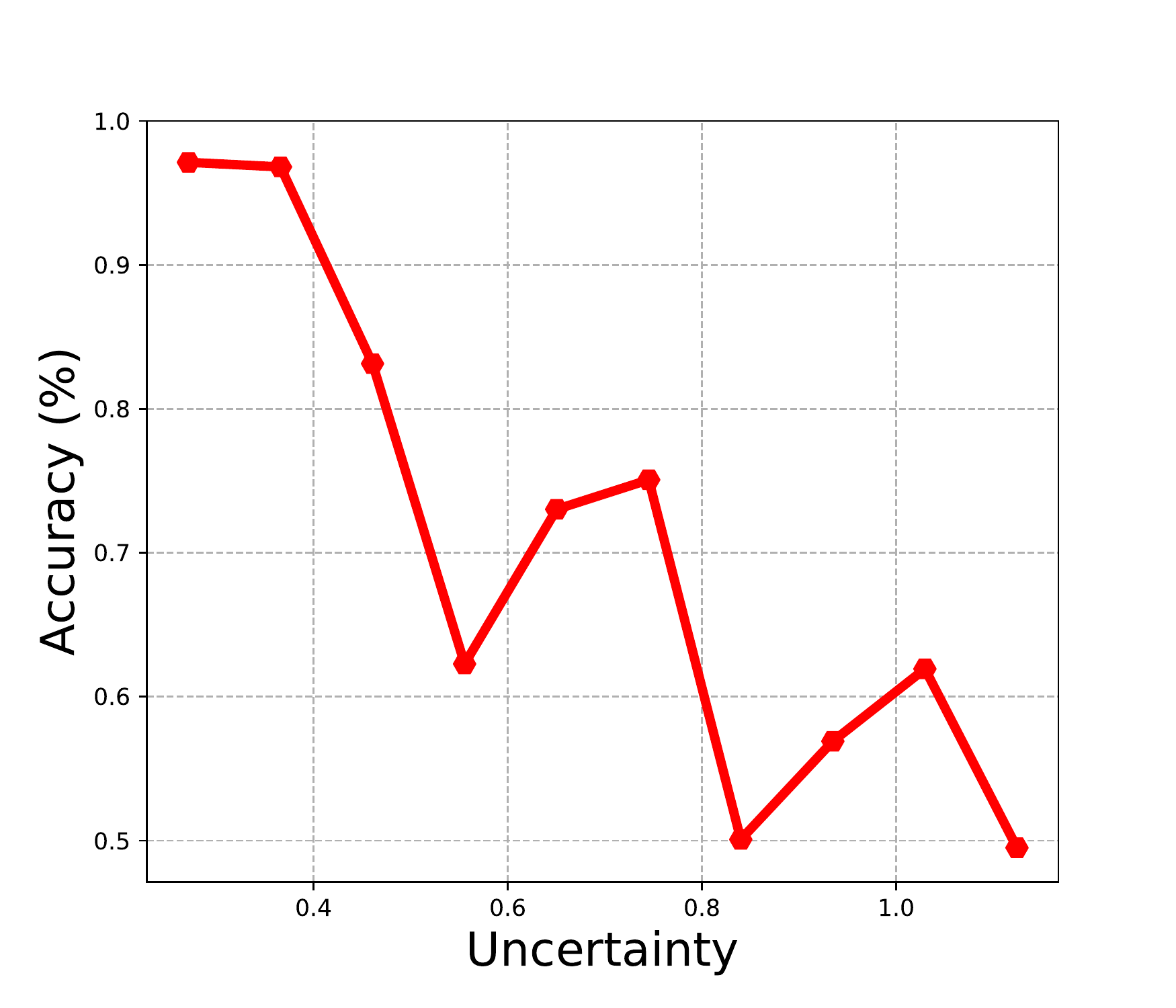}
  }
  \subfigure[MAE vs uncertainty on Adience dataset]{
    \label{fig:corre:adience2}
    \includegraphics[width=0.32\linewidth]{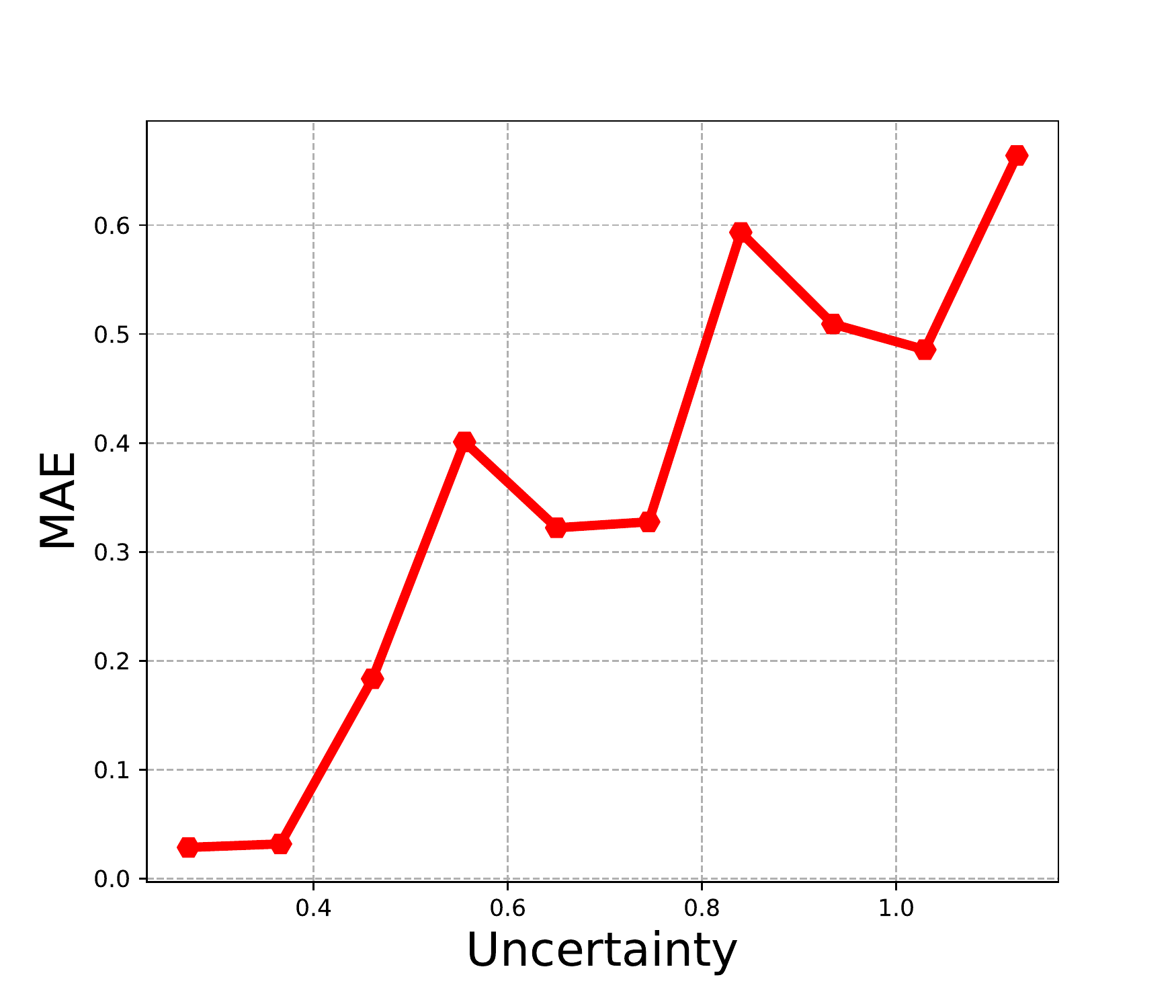}
  }
  \caption{Correlations between the uncertainty score and different performance metrics on two corrupted test sets. To quantity the correlation, we adopt the Kendall's tau correlations~\cite{kendall1938new}. For the curves of three sub-figures, the results are 0.956, -0.733, and 0.778, respectively.}
  \label{fig:corre}
  \vspace{-0.5cm}
\end{figure*}

\textbf{Comparison with Other Methods:} Table \ref{table:SOTA:MORPH}  shows the experimental results on the MORPH II database. We list the results of three different regression methods and their corresponding  probabilistic ordinal embeddings versions. The D-Regression denotes the direct regression method. Equipped with the proposed POE, the direct regression approach, classification based method, and ranking based solution achieve an MAE of 2.50, 2.35, and 2.43 respectively, which outperform the versions of deterministic embeddings by a large margin. By modeling the data uncertainty in the embedding space, our method essentially learns a complex probability model for regression results, which leads to consistent performance improvement. We also compare our method with some state-of-the-art methods~\cite{tan2018efficient,wen2020adaptive} which are specially designed for age estimation. For example, DRFs~\cite{shen2018deep} extended the deep neural decision forests~\cite{kontschieder2015deep} to the regression problem and proposed deep regression forests to learn a Gaussian mixture model. AVDL~\cite{wen2020adaptive} learned an adaptive variance based distribution for labels with a meta-learning framework and achieved excellent results. Compared with these advanced approaches, POE based methods achieve superior or competitive results, which illustrates the effectiveness of our method.

We also conducted experiments on the Adience face database. Table \ref{table:SOTA:Adience} tabulates the results. We observe that the proposed  probabilistic ordinal embeddings consistently improve the performance of three commonly used regression methods. For example, our method improves the ranking based regression solution by 3.6\% for accuracy and 0.06 years for MAE. With the proposed POEs, the classification based method attains an accuracy of 60.5\% and an MAE of 0.47. We see that it also outperforms the previous state-of-the-art methods, which shows the superiority of our probabilistic ordinal embeddings.

\textbf{Uncertainty Estimation:} Knowing what a model does not know is critical for many real-world applications. We aim to estimate when an input image can be reliably predicted or not. Since the data uncertainty is modeled in the diagonal covariance $\bm{\Sigma}(\bm{x})$, we give a per-exemplar uncertainty score using the harmonic mean of the predicted variance: $D/\sum_{j=1}^D (\bm{\sigma}^j)^{-1}$.

To validate the utility of our uncertainty score, we conducted experiments on the MOPRH II database and the Adience face database. We first construct corrupted test sets for both databases using gaussian blur with various radiuses. For each test set, we sort all samples based on the defined uncertainty scores which are calculated with our best model. Then we divide all sorted test samples into 10 bins. For the samples in a bin, we calculate the mean absolute error to show the correlations. For the Adience face database, we also employ the classification accuracy metric. We plot the correlations between performance metrics and our uncertainty score in Figure \ref{fig:corre}. We see that the proposed uncertainty score correlates well with MAE on the MORPH II database, which demonstrates that our uncertainty score can be regarded as an indicator of the reliability of model predictions. We observe the same trend with some jitters on the Adience face database, which further verifies the utility of our uncertainty score.

\begin{table*}[t]
\caption{Results on the Image Aesthetics dataset. We report the accuracy and MAE for each nominal category.}
\vspace {0.2cm}
\label{table:SOTA:Aesthetics}
\renewcommand\tabcolsep{5pt}
\centering
\begin{tabular}{lcccccccccc}
\toprule
\multirow{2}*{Methods}  & \multicolumn{5}{c}{Accuracy(\%) - higher is better} & \multicolumn{5}{c}{MAE - lower is better}  \\
\cmidrule(r){2-6}  \cmidrule(r){7-11}
~ & Nature & Animal & Urban & People & Overall & Nature & Animal & Urban & People & Overall \\
\midrule
CNNPOR   \cite{liu2018constrained} & 71.86 & 69.32 & 69.09 & 69.94 & 70.05 & 0.294 & 0.322 & 0.325 & 0.321 & 0.316  \\
SORD \cite{diaz2019soft} & 73.59 & 70.29 & \textbf{73.25} & 70.59 & 72.03 & 0.271 & 0.308 & \textbf{0.276} & 0.309 & 0.290 \\
\midrule
D-Regression & 71.52 & 70.72 & 71.22 & 69.72 & 70.80 & 0.378 & 0.397 & 0.387 & 0.400 & 0.390  \\
D-Regression  + POE  & 73.44 & 70.44 & 72.26 & 71.02 & 71.79 & 0.360 & 0.386 & 0.376 & 0.385 & 0.376 \\
Classification~\cite{liu2018constrained} & 70.97 & 68.02 & 68.19 & 71.63 & 69.45 & 0.305 & 0.342 & 0.374 & 0.412 & 0.376  \\
Classification + POE & 73.62 & \textbf{71.14} & 72.78 & \textbf{72.22} & \textbf{72.44} & 0.273 & \textbf{0.299} & 0.281 & \textbf{0.293} & \textbf{0.287} \\
Ranking~\cite{niu2016ordinal} & 69.81 & 69.10 & 66.49 & 70.44 & 68.96 & 0.313 & 0.331 & 0.349 & 0.312 & 0.326  \\
Ranking + POE & \textbf{74.40} & 71.12 & 72.22 & 71.58 & 72.33 & \textbf{0.268} & 0.301 & 0.288 & 0.298 & 0.289   \\
\bottomrule
\end{tabular}
\vspace {-0.5cm}
\end{table*}

\subsection{Image Aesthetics Assessment}
\textbf{Dataset:} The Aesthetics dataset ~\cite{schifanella2015image} is used to estimate the aesthetic grades, where 15,687 Flickr image URLs are provided and 13,929 images are available. Four categories are considered in this dataset: nature, animal, urban, and people. The 5-point absolute category rating scale is employed to evaluate the photographic quality: unacceptable, flawed, ordinary, professional, and exceptional. Each image receives at least five judgments from different graders. The ground truth is defined as the median rank. We randomly divide all images into three non-overlapped subsets: 75\% for training, 5\% for validation, and 20\% for testing. Following ~\cite{liu2018constrained,diaz2019soft}, we perform five-fold cross-validation for a fair comparison.

\textbf{Experimental Settings:} Since most previous works~\cite{liu2018constrained,liu2019probabilistic,diaz2019soft} employed the ImageNet pre-trained VGG-16 as the backbone, we also adopted it to extract image features. For optimization, we used Adam optimizer with a batch size of 32. The learning rate was set to 0.001 for the last layer and 0.0001 for other layers. For data preprocessing, we first resized all training image into 256 $\times$ 256, and then randomly cropped them to 224 $\times$ 224 patches. Random horizontal flipping was further performed for data augmentation. Both classification accuracy and MAE were utilized as the evaluation metrics. For the direct regression method and ranking based method, we obtained the predicted class with the same strategy used in age estimation databases.

\textbf{Results:} Table \ref{table:SOTA:Aesthetics} summarizes the results on the Image Aesthetics dataset for each nominal category. We observe that our probabilistic ordinal embeddings significantly boost existing regression methods. For example, the POE version of the ranking based method achieves an overall accuracy of 72.33\% and an overall MAE of 0.289, outperforming the original ranking method by 3.37\% for accuracy and 0.037 for MAE. The consistent performance improvements across all nominal categories demonstrate the effectiveness of our approach. What's more, the classification based method with the proposed probabilistic ordinal embeddings achieves the state-of-the-art results, which further shows the importance of data uncertainty modeling.

\begin{table}[t]
\caption{Results of our models and state-of-the-art methods on the historical image benchmark. We report the accuracy and MAE.}
\vspace {0.2cm}
\label{table:SOTA:historical}
\centering
\begin{tabular}{lcc}
\toprule
Methods  & Accuracy(\%) & MAE  \\
\midrule
Palermo \emph{et al.} \cite{palermo2012dating} & 44.92 $\pm$ 3.69 & 0.93 $\pm$ 0.08 \\
CNNPOR   \cite{liu2018constrained} & 50.12 $\pm$ 2.65 & 0.82 $\pm$ 0.05   \\
GP-DNNOR \cite{liu2019probabilistic} & 46.60 $\pm$ 2.98 & 0.76 $\pm$ 0.05  \\
\midrule
D-Regression & 42.24 $\pm$ 2.91 & 0.79 $\pm$ 0.03   \\
D-Regression  + POE & 46.04  $\pm$ 2.87 & 0.72 $\pm$ 0.02   \\
Classification~\cite{liu2018constrained} & 48.94 $\pm$ 2.54 & 0.89 $\pm$ 0.06   \\
Classification + POE & \textbf{54.68 $\pm$ 3.21} & 0.67 $\pm$ 0.04   \\
Ranking~\cite{niu2016ordinal} & 44.67 $\pm$ 4.24 & 0.81 $\pm$ 0.06   \\
Ranking + POE & 50.01 $\pm$ 4.12 & \textbf{0.66 $\pm$ 0.05}  \\
\bottomrule
\end{tabular}
\vspace {-0.5cm}
\end{table}

\subsection{Historical Image Dating}
\textbf{Dataset:} The historical color image dataset~\cite{palermo2012dating} is collected for the task of automatically estimating the age of historical color photos. Each image is annotated with its associated decade, where five decades from the 1930s to 1970s are considered. There are 265 images for each category. Following the setting used in \cite{liu2018constrained,liu2019probabilistic}, the standard train/val/test split uses 210 images for training, 5 images for validation, and 50 images for testing in each decade. Therefore, the total size of training, validation, and testing sets are 1,050, 25, and 250 images, respectively. Since this benchmark is small scale, we adopt ten-fold cross-validation as used in \cite{palermo2012dating,liu2018constrained} and report the mean values of the results.

\textbf{Experimental Settings:} All experiments in this dataset utilized VGG-16 as the backbone, which was initialized with the ImageNet pre-trained weights for a fair comparison. We set the initial learning rate to 0.0001 except for the last fully-connected layer, which used a higher learning rate of 0.001 to accelerate convergence. We trained the network for 50 epochs with Adam optimizer. To avoid overfitting, data augmentation was applied in our experiments. Concretely, we performed random horizontal flipping and random cropping for each training image. The prediction was obtained with a central crop during testing.

\textbf{Results:} We show the results on the historical color image dataset in Table \ref{table:SOTA:historical}. It is observed that the performance of three different regression approaches is greatly improved by the proposed probabilistic ordinal embeddings. Specifically, the classification based method with our POEs achieves an accuracy of 54.68\% and an MAE of 0.67, which outperforms the deterministic embeddings version by 5.74\% for accuracy and 0.22 for MAE. Combined with POEs, the ranking based method achieves the lowest MAE of 0.66 and outperforms the state-of-the-art method CNNPOR~\cite{liu2018constrained} by 0.1 years, which illustrates the superiority of our proposed probabilistic ordinal embeddings.

%
%
%

\section{Conclusion}

In this paper, we have presented the probabilistic ordinal embeddings for uncertainty-aware regression. Unlike the existing methods that only consider the direct regression solution and model the data uncertainty in the output space with a certain family of probabilistic distributions, our method models the uncertainty within the popular regression methods. We treat each data as a multivariate Gaussian distribution in the embedding space and apply the ordinal distribution constraint to enforce the ordinal relations in the latent space. Our method provides a per-exemplar uncertainty score, which can be used to measure the reliability of model predictions. Extensive experiment results on three visual tasks including age estimation, image aesthetics assessment, and historical image dating demonstrate that our method significantly boosts the present-day regression methods and achieves state-of-the-art performance.

\section*{Acknowledgement}
This work was supported in part by the National Natural Science Foundation of China under Grant 61822603, Grant U1813218, Grant U1713214, in part by Beijing Academy of Artificial Intelligence (BAAI), and in part by a grant from the Institute for Guo Qiang, Tsinghua University.

{\small
\bibliographystyle{ieee_fullname}
\bibliography{egbib}
}

\newpage
\appendix
\begin{center}
\noindent{\textbf{\large{Supplementary Materials}}}
\end{center}

\section{Parameters Discussion}

In this section, we investigate the influence of different weights of ordinal constraint loss $\alpha$ and VIB loss $\beta$ on the MORPH II dataset. In previous experiments, we fixed the $\alpha$ to 1e-4, $\beta$ to 1e-5. To see the effect of these trade-off hyper-parameters, we freeze one parameter as default and tweak the other one. Table~\ref{table:ablation:alpha} shows the comparison results with different $\alpha$ values on the MORPH database. Our method achieves the best performance when $\alpha$ is set to 1e-4. Therefore we set $\alpha$ to 1e-4 in other experiments.  Table~\ref{table:ablation:beta} illustrates the results with varied $\beta$ values on the MORPH database. The best result is achieved when $\beta$ is set to 1e-5, which is adopted in the following experiments. We also observe significant performance degradation with large $\alpha$ or $\beta$ weights (larger than 1e-3), which demonstrates the necessity of hyper-parameter tuning.

\begin{table}[htbp]
\caption{Results of ordinal loss with different $\alpha$ values on the MORPH II dataset.}
\vspace {0.2cm}
\label{table:ablation:alpha}
\centering
\begin{tabular}{lcccccc}
\toprule
$\alpha$ & 1e-2 & 1e-3 & 1e-4 & 1e-5 & 1e-6 & 1e-7\\
\midrule
MAE & 3.89 & 2.42 & \textbf{2.35} & 2.39 & 2.38 & 2.40\\
\bottomrule
\end{tabular}
\vspace{-0.5cm}
\end{table}

\begin{table}[htbp]
\caption{Results of VIB loss with different $\beta$ values on the MORPH II dataset.}
\vspace {0.2cm}
\label{table:ablation:beta}
\centering
\begin{tabular}{lcccccc}
\toprule
$\beta$ & 1e-2 & 1e-3 & 1e-4 & 1e-5 & 1e-6 & 1e-7\\
\midrule
MAE & 8.36 & 2.39 & 2.39 & \textbf{2.35} & 2.37 & 2.37\\
\bottomrule
\end{tabular}
\vspace{-0.5cm}
\end{table}

\section{Distributions of Uncertainty}
The harmonic mean of the predicted variance $\bm{\sigma}$ is employed as the approximated measurement of the estimated uncertainty. To further demonstrate the utility of this metric,  We visualize the distributions of the learned uncertainty on the corrupted MORPH II test set using Gaussian blur with three different radii. The results are shown in Figure 1.  With different Gaussian blur radii (each equal to 0, 5, 10), the learned uncertainty increases while the image quality degrades. As one can see, the distributions "move" to the right by a large margin as the quality degradation increases in the following order: radius=0 $<$ radius=5 $<$ radius=10.

\begin{figure}[h]
  \centering
  \label{fig:uncertainty}
  \includegraphics[width=1.0\linewidth]{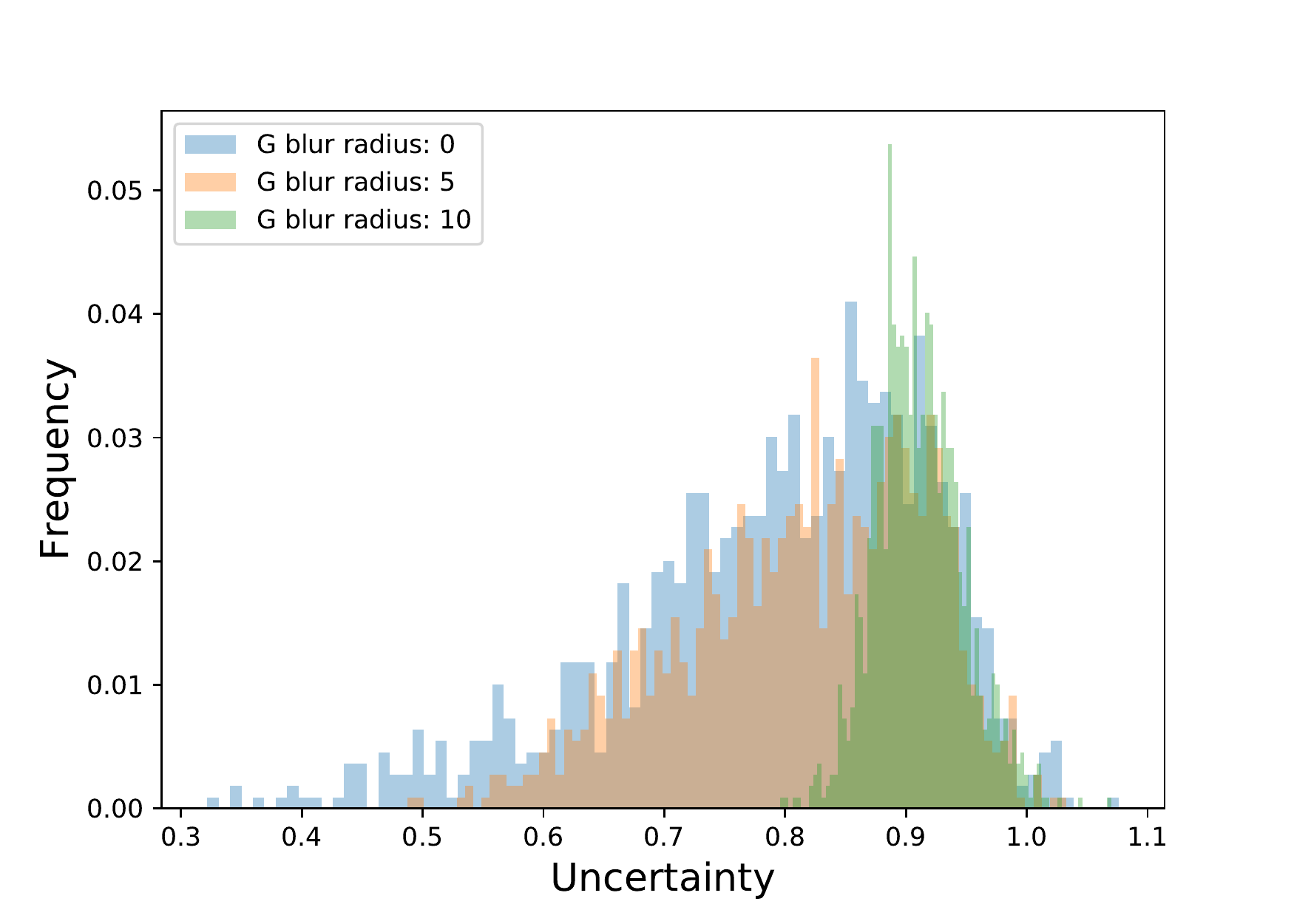}
  \caption{Uncertainty distributions on the corrupted MORPH II test set.}
\end{figure}

\begin{figure*}[t]
  \centering
  \includegraphics[width=1.0\linewidth]{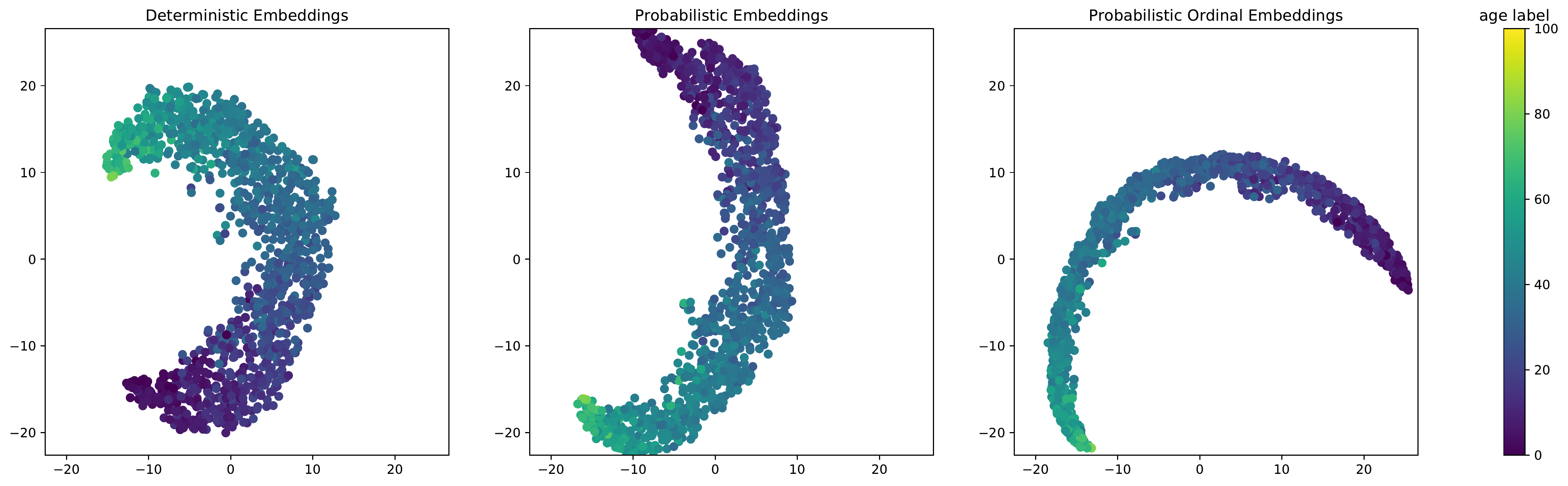}
  \caption{Visualization of feature embeddings with the classification based method on the MORPH II test set.}
  \label{figcls}
\end{figure*}

\section{Triangle Inequality Issue}
The potential problem of symmetric KL divergence is that it is not a strict distance metric due to the triangle inequality issue. However, it is widely used to measure the difference of probability distributions in practice and achieves great success in many areas such as VAE. We also provide the 2-Wasserstein distance in our paper and find that both of them work well. We present more results on other tasks in Table \ref{table:metric} and observe the same conclusion. Users can choose the suitable metric according to their needs.

\begin{figure*}[t]
  \centering
  \includegraphics[width=1.0\linewidth]{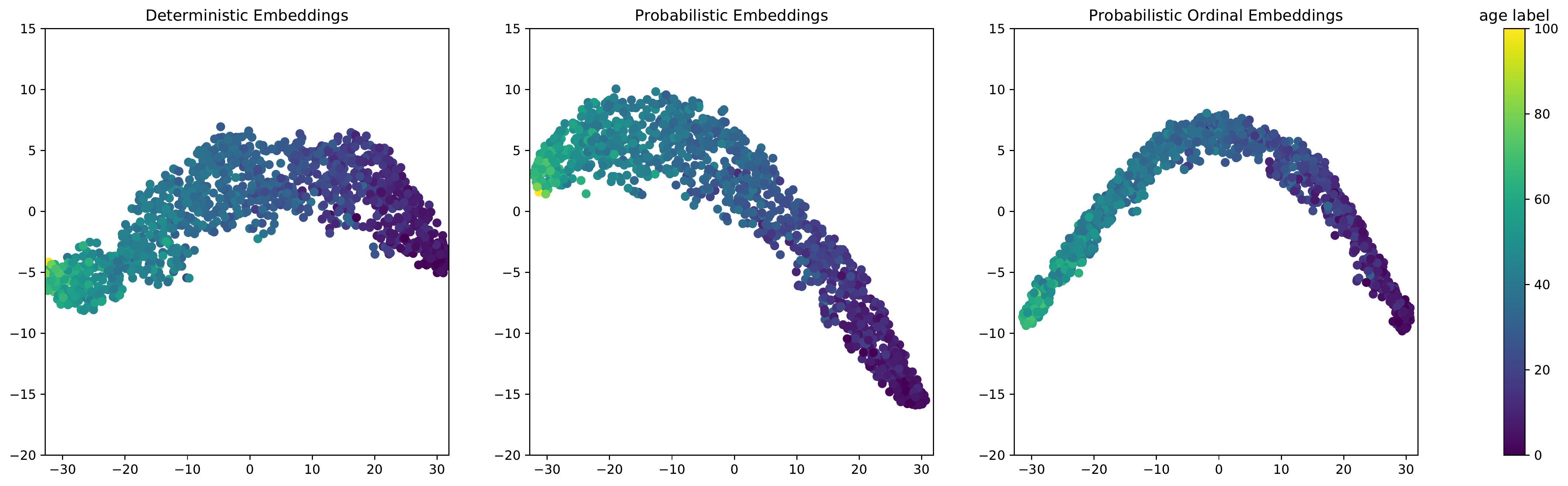}
  \caption{Visualization of feature embeddings with the ranking based method on the MORPH II test set.}
  \label{figrank}
\end{figure*}

\begin{table}[h]
\caption{More results with different metrics. 2-W for 2-Wasserstein distance and S-KL for symmetric KL divergence.}
\vspace{-0.5cm}
\begin{center}
\renewcommand\tabcolsep{4pt}
\begin{tabular}{l|c|c|c|c}
\hline
\multirow{2}*{Metric} & \multicolumn{2}{c|}{Adience}   &
\multicolumn{2}{c}{Historical}
 \\
 \cline{2-5}
~ &  Acc (\%) & MAE & ACC (\%) & MAE  \\
\hline
2-W & 61.7$\pm$4.7 &  0.45$\pm$0.07 & 51.88$\pm$1.98  & 0.72$\pm$0.03   \\
\hline
S-KL & 60.5$\pm$4.4 & 0.47$\pm$0.06 & 54.68$\pm$3.21  & 0.67$\pm$0.04 \\
\hline
\end{tabular}
\end{center}
\label{table:metric}
\vspace{-0.5cm}
\end{table}


\section{Ordinal Information}


To validate that the proposed method can preserve the ordinal information in the embedding space, we conducted  quantitative and qualitative evaluations. For quantitative results, we count the proportion of triplets that violate the ordinal constraint in the embedding space on the test set. The results on the MORPH II test set are presented in Table \ref{table:quantitative}. We see that our POE better preserves the ordinal information in the embeddings.
For qualitative results, we illustrate the learned features of deterministic embeddings, probabilistic embeddings, and probabilistic ordinal embeddings on the MORPH II test set. Figure~\ref{figcls} shows the results with the classification based method and Figure~\ref{figrank} shows the results with the ranking based method. When using t-SNE, we set the perplexity=100 and fix the initial state. We see that compared with the deterministic embeddings and probabilistic embeddings, POEs learned more compact and ordered feature embeddings, which validates that the ordinal constraint in the target space is well preserved in the learned embedding space.

\begin{table}[h]
\caption{Quantitative results with different methods. D-E for Deterministic Embeddings, P-E for Probabilistic Embeddings, POE for Probabilistic Ordinal Embeddings (our method). }
\begin{center}
\renewcommand\tabcolsep{5pt}
\begin{tabular}{c|c|c|c|c|c|c}
\hline
\multirow{2}*{Method} & \multicolumn{3}{c|}{Classification based}   &
\multicolumn{3}{c}{Ranking based}
 \\
\cline{2-7}
~ &  D-E & P-E & POE & D-E & P-E & POE  \\
\hline
\% & 28.71 &  30.44 & \textbf{15.58}  & 19.93 & 18.87 & \textbf{15.99}   \\
\hline
\end{tabular}
\end{center}
\label{table:quantitative}
\end{table}

\section{Standard Deviation}
Following most previous works, we did not report the standard deviations for the experiments on the MORPH II dataset. Now we list the standard deviations according to the order in Table 3 of our main paper: 0.02, 0.06, 0.02, 0.01, 0.01, 0.03, 0.06, 0.03, 0.02, 0.01. The standard deviations of all six methods in Table 4 of our main paper are as follows: 0.03, 0.01, 0.02, 0.01, 0.03, 0.01. We observe very small standard deviations, indicating that the original conclusions still hold.

\section{The Selection of $T$}
Monte-Carlo sampling is \textbf{ONLY} used during \textbf{training}.
To determine the number $T$ of samples, we show the comparisons of multiply-accumulate operations (MACs) and  performance with different sample numbers $T$ in Table \ref{table:samples}.  We set $T$ = 50 for a good trade-off in our experiments. Note that the extra computation costs are \textbf{tiny ($\sim$0.1\%)}.

\begin{table}[h]
\caption{The comparison of  MACs and performance  with different numbers of samples $T$ on the MORPH II dataset. T = 0 indicates deterministic embeddings. }
\vspace{-0.5cm}
\begin{center}
\renewcommand\tabcolsep{5pt}
\begin{tabular}{l|c|c|c|c|c}
\hline
$T$ &  0  & 10 & 50 & 100 & 200  \\
\hline
MACs (G) & 15.497 & 15.501  & 15.517 & 15.538 & 15.579  \\
\hline
MAE & 2.64  & 2.43  & 2.35 & 2.36 & 2.34  \\
\hline
\end{tabular}
\end{center}
\label{table:samples}
\end{table}

\section{Online Hard Example Mining}
For a triplet $(\bm{x}_l,\bm{x}_m,\bm{x}_n)$, the relationship is $|\bm{y}_l -\bm{y}_m| < |\bm{y}_l -\bm{y}_n|$, and we aim to constrain the probabilistic embeddings to satisfy  $d(\bm{z}_l,\bm{z}_m) < d(\bm{z}_l,\bm{z}_n)$.
The proposed ordinal distribution constraint involves triplet selection. Generating all possible triplets from $\mathcal{S} = \{(l,m,n)\big| \left| \bm{y}_l - \bm{y}_m \right| < \left| \bm{y}_l - \bm{y}_n \right|\}$ is inefficient since many of them easily fulfill the ordinal distribution constraint and do not contribute to the network training. One way to address this issue is to select hard triplets. In this paper, we adopt a simple online hard example mining strategy to ensure fast convergence. We assume that the triplet $(l,m,n)$ is a hard one when the difference between  $ \left| \bm{y}_l - \bm{y}_m \right|$ and $\left| \bm{y}_l - \bm{y}_n \right|$ is small. For a batch of training data with $N$ samples, we construct $N$ triplets from them. We first set  each sample in the batch as the $anchor$ sample $l$. Then the next adjacent sample in the batch is selected as the second element of the corresponding triplet. The third element is chosen from the remaining $N-2$ samples which is most likely to violate the constraint according to the above assumption. Specifically, the sample that minimizes $\big| \left| \bm{y}_l - \bm{y}_m \right| -  \left| \bm{y}_l - \bm{y}_n \right| \big|$ and satisfies $ \left| \bm{y}_l - \bm{y}_m \right| \neq  \left| \bm{y}_l - \bm{y}_n \right|$ is selected. In this way, we obtain $N$ triplets for each batch. Experimental results show that this strategy can achieve satisfactory results. It is an interesting future work to explore more advanced hard example mining strategies.

\end{document}